Emin Nakilcioğlu, Anisa Rizvanolli and Olaf Rendel

# Workload Forecasting of a Logistic Node Using Bayesian Neural Networks

HICL

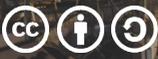

# Workload Forecasting of a Logistic Node Using Bayesian Neural Networks

Emin Nakilcioğlu[1], Anisa Rizvanolli [1] and Olaf Rendel[1]

1 – Fraunhofer Center for Maritime Logistics and Services (CML)

**Purpose:** Traffic volume in empty container depots has been highly volatile due to external factors. Forecasting the expected container truck traffic along with having a dynamic module to foresee the future workload plays a critical role in improving the work efficiency. This paper studies the relevant literature and designs a forecasting model addressing the aforementioned issues.

**Methodology:** The paper develops a forecasting model to predict hourly work and traffic volume of container trucks in an empty container depot using a Bayesian Neural Network based model. Furthermore, the paper experiments with datasets with different characteristics to assess the model's forecasting range for various data sources.

**Findings:** The real data of an empty container depot is utilized to develop a forecasting model and to later verify the capabilities of the model. The findings show the performance validity of the model and provide the groundwork to build an effective traffic and workload planning system for the empty container depot in question.

**Originality:** This paper proposes a Bayesian deep learning-based forecasting model for traffic and workload of an empty container depot using real-world data. This designed and implemented forecasting model offers a solution with which every actor in the container truck transportation benefits from the optimized workload.

Workload Forecasting of a Logistic Node Using Bayesian Neural Networks

# 1 Introduction

International trade has been one of the key factors in the development of world economies, increasing the need for efficient supply chain mechanisms for product and service distribution in global markets (Rodrigue, 2009). "Although the responsiveness of trade to the Gross Domestic Product (GDP) growth has been moderate over the recent years, demand for maritime transport services and seaborne trade volumes continue to be shaped by global economic growth and the need to carry merchandise trade" (UNCTAD, 2015). In this respect, inland transport of cargo and empty containers plays an essential role in the efficiency of global supply chains.

Empty container depots function as a central place where the shipping and logistics companies can hold their containers until the reuse of containers for the next shipment.

Short term load forecasting which refers to forecasting up to 1 week serves as a practical tool for container terminals and depots to plan the day to day operation (Widhyohadi, 2013). Accurate forecasts of the work or system load on an hourly basis from one day to a week ahead help the container depot operator to accomplish a variety of tasks like economic scheduling of worker capacity, scheduling the necessary handling equipment, etc. In particular, forecasting the peak demand is considered essential as the worker capacity must be adequate in those times to appropriately and efficiently handle the incoming and outgoing traffic of container trucks. Since such forecasting applications lead to increase in the security of operation conditions and cost savings, numerous techniques have been utilized to improve the short-term load forecasting (Hippert, et al., 2001).

Though statistical approaches, such as autoregressive modeling, are a long-time staple and still widely used for forecasting applications (Moghram, and Rahman, 1989), significant advancements in the field of deep learning over the recent years has brought deep learning based forecasting techniques into the spotlight. Today deep learning dominates most forecasting applications and provides state-of-the-art performance (Bim, et al, 2021 and Nassif, et al., 2021).

Although the existing research has successfully demonstrated the superior performance of deep learning on forecasting tasks, inherently, most of the studies are actually based on deterministic models, which lack the ability to capture uncertainty. A new probabilistic deep learning model, the concept of Bayesian deep learning (BDL), which enables a deep learning framework to model uncertainty, is becoming increasingly prevalent in computer vision, natural language processing, medical diagnostics, and autonomous driving (Gal, 2016). BDL exhibits the benefits of uncertainty representation, understanding generalization and reliable prediction; leading to a more interpretable deep neural network through the lens of probability theory. In this paper, a novel probabilistic framework based on BDL for forecasting the workload of an empty container depot is proposed. In order to ascertain whether the forecasting model will help empty container depots improve their current working standards in terms of truck handling operations, the model's performance is examined in different setups and conditions. This approach also captures the aleatoric uncertainty of forecasting data and models. Every hour, the proposed framework automatically produces the hourly workload prediction for the upcoming 5 working days. The primary database for the experimental setup is provided by HCS-Depot which is privately managed specialist company for the repair and stock holding of empty containers in the port of Hamburg. A BDL-based forecasting model is trained using this primary database. In order to examine the capability and forecasting performance of the model, a comparative study between the BDL-based model and another forecasting model is performed. The other model that serves as a benchmark model in this experimental setup was developed by Fraunhofer Center for Maritime Logistics and Services (CML) in the scope of Project LILIE where Fraunhofer CML was partnered up with HCS-Depot to develop a forecasting model by using the multilayer perceptron (MLP) approach (Rendel, John and Karnbach, 2018). Compared with the benchmark model, the proposed approach outperforms the conventional MLP approach and the results show the importance of probabilistic approach over deterministic approach. In addition to this initial comparative study, another performance evaluation between the BDL-based forecasting models which are trained using another complimentary datasets is carried out.

The remainder of this paper is structured as follows; Section 2 introduces current deep learning methods used in the context of short-term load forecasting and provides a brief overview of relevant work as well as detailed insight to Bayesian neural networks. The use case "workload forecasting for an empty container Depot" is covered in Section 3 with individual subsections, giving an overview of the workload at the empty container depot of HCS-Depot and describing every dataset used during the model training, the developed model architecture as well as experimental setup and achieved results. Moreover, the experimental results along with the limitations of the study is discussed in this section. Concluding remarks and possibilities for future research are given in Section 4.

## 2 Deep Learning for Forecasting

Over the past few years, artificial intelligence and in particular machine learning have seen a significant increase in terms of its usage. Machine learning consists of a number of different algorithms that have been developed to learn correlations through pattern recognition in datasets and use these correlations to make predictions for new, previously unknown data (Nelli, 2018). Today, supervised learning is considered the mainstream subfield of machine learning. Main element of the supervised learning is a large number of labeled data which is a combination of input data and desired output. The labeled data is automatically processed to learn the statistical correlations that encapsulate the relation between input and output. Eventually, these relationships can be used in terms of decision rules when predicting the corresponding output for a given input (Müller and Guido, 2017). More recently, deep learning was introduced as a new subfield in machine learning.

Machine learning algorithms use labeled data to generate predictions. In other words, particular features are defined from the input data for the model and organized into tables, which indicates that input data generally goes through some pre-processing to achieve a structured format. Some of these data pre-processing is eliminated by deep learning algorithms. The main difference with deep learning is that its algorithms are able to ingest and process unstructured data, such as text data and images. Thus, the feature extraction process is automated and some of the dependency on human experts was removed.

The performance of deep learning models has been significantly exceeding the performance of classical machine learning algorithms in various supervised learning problems. Accelerated by ever-expanding computing power and data volumes, deep learning has therefore enabled notable breakthroughs in applications where text, image or sound data is being processed (Le Cun and Bengio 1995, Goodfellow, et al., 2016 and Le Cun, et al., 2015).

In modern times, forecasting has been a long-time research and application topic among researchers and engineers from different research areas and industries (Goldfarb, et.al, 2005 and The New York Times, 1984). It has a wide range of application areas from time-series forecasting (Lim, et al., 2021 and Deng, et al., 2022) and cloud computing workload forecasting (Masdari and Khoshnevis, 2020) to water quality forecasting for prawn ponds (Dabrowski, et al., 2022) and weather forecasting (Garg, et al., 2022.). As deep learning algorithms and applications started to become readily available and feasible due to worldwide academical and industrial interest along with ever-increasing and affordable computational power getting to a level that is more available for the masses, forecasting has become one of the trendy subfields of machine and deep learning. In the scope of this paper, literature on short-term load forecasting techniques were further investigated and analyzed in detail.

Despite of the increasing trend towards deep learning-based forecasting techniques, there has not been many research studies in the area of short-term load forecasting where the effects of statistical and deterministic approaches were investigated. Cao et al. (2015) adopted an autoregressive integrated moving average (ARIMA) model and similar-day method for intraday load forecasting for electric power companies in China. The study demonstrated that ARIMA performs better in ordinary days while in unordinary days, similar-day yields better results. In another study, Long Short-Term Memory (LSTM) based Recurrent Neural component was used to perform one day ahead hourly load forecasting (He, 2017). The idea explained in the paper was to make use of different types of neural network components to model different types of factors that may impact load consumption. They borrowed the approach in modern image recognition introduced by

Workload Forecasting of a Logistic Node Using Bayesian Neural Networks

Szegedy, et al. (2015) and used multiple Convolutional Neural Network (CNN) components to learn rich feature representation from historical load series. Following that, the variability and dynamics in historical loading were modelled with the LSTM based Recurrent Neural component. For other features such as temperature and holidays, dense (feed-forward) component was employed to project these features into vector representations. As the final step, all learned features were linked together through dense layers to predict load value. Recently, some successful examples on load forecasting were accomplished using k-nearest neighbor (KNN) algorithm whose efficiency was concluded to be the dominant advantage (Zhang, et al., 2016 and Al-Qahtani, et al., 2013)

## 2.1 Bayesian Neural Networks

Deep learning has demonstrated state-of-the-art performance in a vast number of tasks; however, as Ghahramani (2016) has illustrated in his paper, it still suffers from a series of limitations that need to be investigated and resolved such as requiring vast amount of data, i.e. very data hungry, no representation of uncertainty and compute-intensive training and deployment. The probabilistic approach of Bayesian neural networks (BNNs) addresses these challenges and offers an efficient solution. In this part, the benefits and rationale of employing BNNs to conduct workload forecasting is qualitatively explained along with establishing the mathematical derivation employed during the development of the forecasting model.

BNNs are inherently a probabilistic model which employs a deep learning model to represent uncertainty. Unlike traditional neural networks where their network parameters are fixed once trained, network parameters (weights and biases) in BNNs are expressed as conditional probabilities. To generate its results, the Bayesian model directly samples from its parameters instead of noise addition to the output or setup of multiple input scenarios which are the approaches employed in traditional neural networks. Thus, the Bayesian model is fundamentally probabilistic rather than deterministic in nature.

Contrary to traditional deep learning methods, most of the existing Bayesian deep learning approaches can capture uncertainties that can be observed in the data and also in the model characteristics (Gal, 2016). Uncertainties with which Bayesian models deal can be grouped under two main categories; aleatoric uncertainty (also referred to as stochastic uncertainty) and epistemic uncertainty (also referred to as model uncertainty). Aleatoric uncertainty addresses the noise in the data while uncertainties in model parameters and model structure are examined under epistemic uncertainty. In the proposed Bayesian forecasting model, only aleatoric uncertainty is addressed. This uncertainty is captured by placing a distribution with small variance (Gaussian random noise) over the output, and thus the model is expected to learn the variance in the noise as a function of different inputs (Kendall and Gal , 2017).

Traditional deep neural networks make use of their neurons to memorize the information inside the training data, which signifies that the parameters in traditional neural networks does not possess any physical meaning; thus, their values can be arbitrary. However, BNNs calculate their outputs using Bayesian theory to render the parameters so that the network has the ability to 'feel' certain or uncertain about its result. In particular, the model along with the prediction uncertainty can be calibrated through Bayesian deep learning approach to obtain smart systems that know exactly what they do not know.

The issue of limited amount of data offered by the real-world tasks is another matter that conventional deep learning systems cannot address since the extremely high or low model complexity will lead to the issues of overfitting or poor performance, respectively. However, it was observed that in the case of BNNs, less data is needed to make accurate forecasting (Sun et al., 2019, Kendall and Gal, 2017). BNNs can effectively address the overfitting problem via integrating prior knowledge into learning systems which imposes a prior on hidden units or neural network parameters, even with insufficient datasets. Particularly, BNNs enable the network to gain automatic model complexity control and structure learning through the benefits of the built-in implicit regularization (Wang and Yeung, 2016).

Sun et al. (2019) presented a new probabilistic day-ahead net load forecasting technique which captures both aleatoric uncertainty and epistemic uncertainty using Bayesian deep learning. The performance of the net load forecasting was improved by considering residential rooftop photovoltaic (PV) outputs as the input features and exploiting the clustering in subprofiles. With high PV visibility and subprofile clustering, the proposed scheme proved its efficacy.

The literature research has shown that although there has been a number of research where the performance of BNN based load forecasting models was evaluated and compared to previous methods, developing a BNN based forecasting model that focuses on the workload of an empty container depot or of a similar working environment is yet to be realized.

## 3    Use Case "Workload Forecasting for an Empty Container Depot"

Container trucks arriving at the empty container depot undergo certain operations, which are referred to as handling process in this paper. Figure 1 illustrates the handling process of the container trucks at the empty container depot. Once a container truck arrives at the gate of the depot, a corresponding gate-in timestamp for the container truck is generated. Following that, trucks awaits until their documents are checked. Depending on the trucks' main task, they are then moved either to the loading or unloading area. After the loading and unloading operation are complete, the trucks leave the empty container depot as the corresponding gate-out timestamp is generated. The entire process between a truck's arrival and departure is considered as the handling process and the time spent during this process is referred to as handling time.

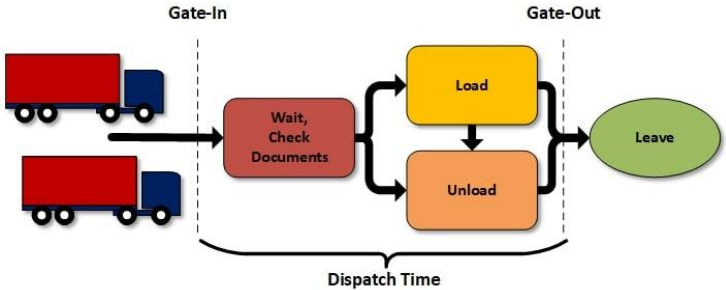

Figure 1: Schematic representation of the workload in the empty container depot

The structure and statistics of the handling process set the framework for developing a hourly forecasting model using Bayesian deep learning approach, as covered in this paper. Accordingly, practical and realistic realization of a forecasting model would require more reliable and stable predictions to be compatible with previously offered solutions. Therefore, the main research question of this paper is whether it is possible to develop a reliable deep learning-based forecasting model which is capable of forecasting the hourly workload of an empty container depot given the recorded data of the depot from the previous years.

The remainder of this section will firstly introduce the datasets utilized throughout the project and then describe the developed and implemented deep learning architecture. Lastly, experimental results of the forecasting model trained with various datasets are presented and discussed along with the study's limitations.

### 3.1.1 Data

In the scope of the project, we have experimented with datasets from different sources in order to observe the individual effects of these datasets on the prediction model and its performance. In this section, the datasets will be further analyzed and explained in detail.

### 3.1.2 Base Dataset

The database that acts as the base dataset for the project was provided by HCS-Depot. The database consists of data records regarding every container truck that is part of the traffic in the empty container depot in question. Certain information about every individual container truck that was handled in the container depot between 2017 and 2021 was recorded by HCS-Depot and these records include gate-in & gate-out, loading & dispatch and gate inspection timestamps of each truck along with additional information regarding the containers the trucks carry. In Table 1, every type of data provided in the database is listed as they were originally named in the database along with their corresponding explanations.

Table 1: Data types provided in the HCS-Depot database and their descriptions.

| Data Type | Description |
| --- | --- |
| **GateInTime** | Timestamp of the truck's arrival at the terminal |
| **GateOutTime** | Timestamp of the truck's departure from the terminal |
| **LoadingTime** | Timestamp of loading or unloading of the individual container |
| **DispatchTime** | Timestamp of dispatch in the interchange |
| **CustomerID** | ID of the customer for whom a container is delivered or collected |
| **Is20Feet** | Defines whether a container was a 20ft container (represented with '1') or a 40ft container (represented with '0') |

| Data Type | Description |
|---|---|
| **IsInbound** | Defines whether a container was delivered (represented with '1') or picked up (represented with '0') |
| **DepotMoveID** | Group key of all containers delivered or picked up on a truck |

As mentioned, the main objective of the paper is to develop a forecasting model that predicts the number of trucks and average handling time in minutes for any given working hour of the upcoming week. Thus, an initial preprocessing step was applied on the original database in order to shape the data according to the purpose of the research. The input data which serves as the base input for the experiments consists of every working hour of a day and the corresponding day of the week while the output data contained hourly truck rate and average truck handling time in the container depot for each given working hour of the week. A snippet of the base training data is shown in Table 2.

Table 2: A snippet from the base training dataset

| | Input | | Output | |
|---|---|---|---|---|
| **Date** | Hour | Day of the Week | Truck Rate | Handling Time |
| **…** | … | … | … | … |
| **2021-06-23 18:00:00** | 18 | 2 | 28 | 16,76 |
| **2021-06-23 19:00:00** | 19 | 2 | 12 | 17,11 |

Workload Forecasting of a Logistic Node Using Bayesian Neural Networks

|                      | Input |   |   | Output |
| --- | --- | --- | --- | --- |
| **2021-06-23 20:00:00** | 20 | 2 | 2 | 7,41 |
| **2021-06-24 05:00:00** | 5 | 3 | 18 | 40,28 |
| **2021-06-24 06:00:00** | 6 | 3 | 41 | 30,31 |
| **…** | … | … | … | … |

### 3.1.3 Trucker Appointment Data

HCS-Depot makes use of an appointment system where the truck drivers give their expected arrival time in advance. Hourly expected truck rate was derived from this data and appended onto the base input data in order to observe whether data from the trucker appointment system would have any influence on the model's prediction performance. Since the forecasting model was expected to predict the truck rate and average handling time for the entire upcoming week every day, appointments made within the prediction day would not be implemented in the prediction process. Therefore, only the appointments made one day before or earlier in the trucker appointment system was utilized to derive hourly expected truck rate. Afterwards, this data was added to the base input data as an additional column as displayed in Table 3.

Table 3: A snippet of the input dataset with added expected truck rate values

| Date | Hour | Day of the Week | Expected Truck Rate |
| --- | --- | --- | --- |
| **…** | … | … | … |
| **2021-06-23 18:00:00** | 18 | 2 | 5 |
| **2021-06-23 19:00:00** | 19 | 2 | 2 |

| Date | Hour | Day of the Week | Expected Truck Rate |
|---|---|---|---|
| **2021-06-23 20:00:00** | 20 | 2 | 0 |
| **2021-06-23 05:00:00** | 5 | 3 | 17 |
| **2021-06-23 06:00:00** | 6 | 3 | 30 |
| **…** | … | … | … |

### 3.1.4 CAx Data

The Container Availability Index (CAx) is an index that gives information about the import and export moves of full containers around major ports. Interested parties can grab the readily available CAx data to monitor the movement of the containers around provided ports.

CAx values are defined between 0 and 1. A CAx value of 0.5 means that the same number of containers leave and enter a port in the same week. Values greater than 0.5 indicates that more containers have entered the port compared to the containers that have left the same port, whereas values smaller than 0.5 indicate an opposite trend. In addition, 20ft and 40ft containers are represented separately in the CAx data. In Figure 2, a graph corresponding to the CAx values for 40ft containers in the Port of Hamburg between January, 2021 and May, 2022 is shown. The horizontal axis of the graph refers to the weeks of a year, whereas the vertical axis represents the CAx values.

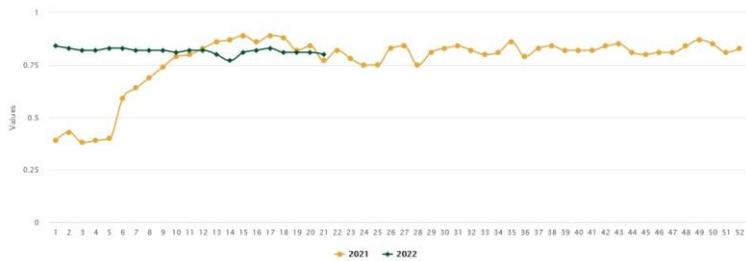

Figure 2: CAx values of 40ft containers in the Port of Hamburg for 2021 and 2022. Source: Container xChange, 2022

Corresponding weekly CAx values were appended to base input data as an additional column.

### 3.1.5 Sailing List

Another additional data source is the sailing list dataset published online by Hamburger Hafen und Logistik AG (HHLA). The dataset provides information about various data of ships in a predefined period (HHLA, 2021). The sailing list dataset serves as a ground data to calculate the corresponding the twenty-foot equivalent unit (TEU) for every ship in the dataset. TEU is an inexact unit of cargo capacity, often used for container ships and container ports. It is based on the volume of a 20-foot-long (6.1 m) intermodal container, a standard-sized metal box which can be easily transferred between different modes of transportation, such as ships, trains and trucks.

The sailing list data provides information regarding the container traffic at the port of Hamburg. As we were achieving to predict the workload of an empty container depot, the amount of time truck drivers spend on the road between the port and the depot was also included in the calculations regarding the gate-in and & gate-out times. After completing the necessary preprocessing of the data, corresponding hourly TEU values were appended to the base input data.

## 3.2 Architecture

Bayesian neural network architecture used to train a forecasting model for the use case of "Workload Forecasting for an Empty Container Depot" consists of an input layer followed by the hidden layers and the output layer, also referred to as predictor.

Hidden layers contain two dense layers. Dense layer is a layer that is deeply connected with its previous layer which means the neurons of the dense layer are connected to every neuron of its previous layer (Dillon, et al., 217). To introduce the aleatoric uncertainty, dense layers are combined with probabilistic layers. With that combination, the mean and covariance matrix of the output is modelled as a function of the input and parameter weights. To achieve that, a multivariate normal distribution was introduced to the dense layer. Through this layer, posterior probability distribution structure, which in this case, due to symmetry, is a multivariate normal distribution where only one half of the covariance matrix is estimated, is declared. Lastly, Kullback–Leibler divergence (KL-divergence) regularizer, which is a regularizer that adds a KL-Divergence penalty to the model's loss function, is added to the output layer (Wang and Ghosh, 2011). The regularizer acts as priori for the output layer.

This novel architecture is developed using TensorFlow which is an interface for expressing machine learning algorithms and an implementation for executing such algorithms (Abadi, et al., 2015). The model also utilized TensorFlow Probability (TFP) library to implement and combine probabilistic models with traditional model components.

## 3.3 Experimental Results

In this section, the structural plan we have followed during the experimentation is explained and corresponding experimental results are presented.

### 3.3.1 Structure of the experiments

In the scope of this paper, various experiments were conducted to analyze the performance of the newly developed forecasting model. The initial experiment in this paper aimed to compare the performance of two different forecasters; the prediction

Workload Forecasting of a Logistic Node Using Bayesian Neural Networks

model proposed in this paper and the prediction model that was previously developed by Fraunhofer CML (Rendel, John and Karnbach, 2018). Both models were subjected to the same input data which is referred to as base input data in this paper and were expected to predict hourly truck rate and average handling time for 4 weeks of August, 2021. As mentioned in 2.1, BNNs was observed to produce more accurate prediction results when less and current data is introduced. To analyze the influence that the time window of the input data has on the model's prediction performance, the input data was split into 5 different sets where the starting date of the input data was differentiated for each set, as shown in Table 4.

Table 4: Timespan of each input data split

|            | Starting Date  | End Date    |
|------------|----------------|-------------|
| **First Set**  | January, 2017  | July, 2021  |
| **Second Set** | January, 2018  | July, 2021  |
| **Third Set**  | January, 2019  | July, 2021  |
| **Fourth Set** | January, 2020  | July, 2021  |
| **Fifth Set**  | January, 2021  | July, 2021  |

Following the initial experimental setup, additional experiments with extended input datasets generated by appending the relevant data extracted from new data sources to the base input data were also carried out. In these experiments, the focus was to observe the effect of these relevant data on the model's prediction performance. The additional data sources provide data only for the year 2021. Therefore, in the secondary experimental setup, the timespan of the input dataset was the same as the fifth set of the primary experimental setup; from January, 2021 to July, 2021.

Since our forecasting model is a probabilistic model, a Monte Carlo experiment was performed to provide the predictions. In particular, every prediction of a sample x results in a different output y, which is why the expectation over many individual predictions has to be calculated. After producing a number of the predictions through iterations, the prediction outputs were averaged out and published as the final predictions by the forecasting model.

### 3.3.2 Results

Two distinct metrics were applied to compare the model's performance with provided datasets; coefficient of determination and mean square error. The coefficient of determination is a statistical measurement that analyses how differences in one variable can be explained by the difference in a secondary variable, when predicting the outcome of a given event. In other words, this coefficient, denoted as $R^2$, assesses the strength of the linear relationship between two variables.

Given that in a dataset with n samples, $y_1, \dots, y_n$ (collectively known as $y_i$) represents the true values, whereas $f_1, \dots, f_n$ (known as $f_i$) being predicted or modeled corresponding values, the residuals can be formulated as in (1).

$$e_i = y_i - f_i \tag{1}$$

The mean of the observed data, denoted with $\bar{y}$, is derived using (2)

$$\bar{y} = \frac{1}{n} \sum_{i=1}^{n} y_i \tag{2}$$

We can measure the variability of the data with two sums of squares formulas;

- The sum of squares of residuals:

$$SS_{res} = \sum_i (y_i - f_i)^2 = \sum e^2 \tag{3}$$

- The total sum of squares (proportional to the variance of the data):

Workload Forecasting of a Logistic Node Using Bayesian Neural Networks

$$SS_{total} = \sum_i (y_i - \bar{y})^2 \tag{4}$$

- Using these two sums of squares formulas, coefficient of determination ($R^2$) can be calculated as follows,

$$R^2 = 1 - \frac{SS_{res}}{SS_{total}} \tag{5}$$

According to the given formula, the highest possible obtainable $R^2$ value is 1, which can be interpreted as that the predicted values exactly match the observed values. $R^2$ value of 0 indicates a baseline model where for each prediction, the mean of observed data ($\bar{y}$) is predicted. As the predictions get worse, $R^2$ value of the model is expected to obtain negative values.

The mean square error (MSE) is defined as the mean of the square of the difference between actual and predicted values. MSE is a commonly used loss function to assess the quality of a predictor. MSE of a predictor is computed as

$$MSE = \frac{1}{n} \sum_{i=1}^{n} (y_i - f_i)^2 \tag{6}$$

where $y_i$ being the i-th observed value and $f_i$ being the i-th predicted value in a dataset with n samples. An MSE of zero is interpreted as that the predictions are generated with perfect accuracy. As the MSE value increases, the performance of the predictor worsens.

Table 5 shows a performance comparison of prediction model developed in Project LILIE and the Bayesian model in terms their MSE rates. Though the prediction performances for forecasting the number of trucks handled hourly in the empty container depot showed no significant difference between the two forecasting models, the proposed Bayesian model yielded a relatively lower error rate for forecasting the average time spent in a given hour for handling the trucks by the depot workers. Especially for the time windows of 2020 and 2021, the Bayesian model performed significantly superior compared to the benchmark model when predicting the average handling time.

Table 5: MSE Values that LILIE and Bayesian model yielded for base input data with different time windows

|  | LILIE | | Bayesian | |
| --- | --- | --- | --- | --- |
| **Training Data Start Date** | Truck Rate | Handling Time | Truck Rate | Handling Time |
| **2017-01-01** | 6.73 | 8.33 | 6.59 | 6.40 |
| **2018-01-01** | 6.39 | 7.68 | 6.50 | 6.40 |
| **2019-01-01** | 6.49 | 8.64 | 6.48 | 6.16 |
| **2020-01-01** | 6.51 | 23.62 | 6.45 | 6.21 |
| **2021-01-01** | 6.50 | 20.43 | 6.29 | 6.20 |

A similar trend was also observed in the coefficient of determination values among the models. Corresponding coefficient of determination comparison is presented in Table 6. Both forecasting models performed similarly in terms of predicting the hourly truck rate. Their coefficient of determination for this prediction laid around 0.7 which is a strong prediction performance, considering that the coefficient of 1 is considered the perfect prediction model. However, the performance difference between the models became highly obvious when predicting the average handling time. Negative coefficient values of the benchmark model indicate poor forecasting performance. On the other hand, a slight increase in forecasting performance of the Bayesian model was observed, as the time window of the input dataset got closer to the present date. These results also support the conclusions regarding the performance of the BNNs when less and more relevant data are fed into the BNNs (Sun et al., 2019, Kendall and Gal, 2017).

Table 6: Coefficient of determination comparison between LILIE-Model and the Bayesian model

|  | LILIE | | Bayesian | |
|---|---|---|---|---|
| **Training Data Start Date** | Truck Rate | Handling Time | Truck Rate | Handling Time |
| **2017-01-01** | 0.72 | -0.65 | 0.71 | 0.29 |
| **2018-01-01** | 0.74 | -0.07 | 0.72 | 0.29 |
| **2019-01-01** | 0.73 | -0.60 | 0.73 | 0.35 |
| **2020-01-01** | 0.74 | -20.43 | 0.73 | 0.33 |
| **2021-01-01** | 0.72 | -18.06 | 0.74 | 0.32 |

The error rate of the Bayesian model for each dataset is presented in Table 7. In this experimental setup, the performance of the Bayesian model with the base dataset served as the benchmark performance. While experiments with CAx dataset and sailing dataset resulted in rather similar performance compared to the benchmark model, the forecasting model trained with data containing trucker appointment system data yielded a superior performance for predicting hourly truck rate. Since the trucker appointment system provides additional relevant information to the model about the number of trucks for the given hour, it was expected that the model would generate more accurate results compared to the benchmark model setup. By training the model with this dataset, the lowest error rate which is approximately 25% lower than the benchmark model's was achieved.

Table 7: MSE Rates of Bayesian model with the experimentation of different datasets.

|  | **Truck Rate** | **Handling Time** |
| --- | --- | --- |
| **KIK-LEE Base Dataset** | 6.29 | 6.20 |
| **Trucker Appointment** | 4.74 | 6.75 |
| **CAx Dataset** | 6.19 | 6.34 |
| **Sailing Dataset** | 6.42 | 6.55 |

Regarding the coefficient of determination, a similar trend was detected as shown in Table 8. When the forecasting model trained using trucker appointment dataset, its performance reached up to the level of 0.85 in terms of its coefficient of determination for truck rate predictions. Though training the model with other datasets resulted in a decrease in the prediction performance for the average handling time, the poorest performance for handling time predictions was achieved by the model trained with trucker appointment dataset. This indicates that the model prioritizes the truck rate correlation during the training.

Table 8: Corresponding coefficient of determination values for different datasets

|  | **Truck Rate** | **Handling Time** |
| --- | --- | --- |
| **KIK-LEE Base Dataset** | 0.74 | 0.32 |
| **Trucker Appointment** | 0.85 | 0.18 |
| **CAx Dataset** | 0.74 | 0.27 |

|  | **Truck Rate** | **Handling Time** |
|---|---|---|
| **Sailing Dataset** | 0.68 | 0.28 |

## 3.4  Limitations and Discussion

As mentioned in 2.1, Bayesian neural networks can employ both aleatoric and epistemic uncertainty. In the scope of this study, only aleatoric uncertainty was modelled in the experiments. Another limitation of the study was a dataset limitation regarding the sailing dataset. HHLA, publisher of the sailing dataset, openly publishes data that only dates back to three months from the current date. Thus, the corresponding dataset employed in the study was consisted of only the sailing data between May, 2021 and August, 2021.

During the comparison experiment between the LILIE-Model and the Bayesian Model, instability in the LILIE-Model was observed after the time window of 2020 when generating predictions for average handling time. The same instable results were obtained even after repeating the experiments for a number of times. It was suspected that the benchmark model, i.e. LILIE-Model, could not perform the learning process or could not approximate any appropriate forecasting function when less data was provided. Furthermore, performance of the Bayesian model has gotten slightly better as the time window of the input dataset has gotten closer to the present date. These results also support the conclusions drawn in previous research papers regarding the performance of the BNNs when less and more relevant data are fed into the BNNs (Sun et al., 2019, Kendall and Gal, 2017).

In the secondary experimental setup, the effects of the additional relevant data provided by the trucker appointment system on the prediction performance were highly visible. Nearly 25% lower error rate for predicting hourly truck rate compared to the benchmark model indicates that providing additional relevant information about the number of trucks that are expected to be handled for the given hour helped the forecasting model generate more accurate predictions. Also, examining the results obtained in the experiment with CAX data, it can be seen that influence of CAx data on the prediction performance was rather insignificant. CAx data is a data recorded weekly, whereas the prediction models are expected to provide hourly forecast. As a result, the CAx value of a week was appended to every hour of this particular week as an additional column in the input dataset. Therefore, the repetitive nature of the CAx data added no further context or relevant information to the training.

On the other hand, the sailing dataset caused poorer performance in terms of both truck rate and handling time predictions. The data time window limitation of this dataset has had its effect on the prediction results and the forecasting model has given the worst prediction performance with the sailing dataset. Though BNNs are expected to give superior results with less and more relevant data, the short time span of the data has prevented the prediction model catch or create any relevant context.

The proposed approach shows promising results for offering a reliable forecasting framework for empty container depots which can be used as the groundwork for fully realized real-life applications. The results also demonstrate that shifting to the probabilistic approach can benefit the forecasting models in terms of reliable predictions and prediction stability.

On the basis of the obtained results, a number of interesting findings invokes possible future researches. Since only the aleatoric uncertainty is considered for the experiments, capturing both the epistemic and the aleatoric uncertainty together serves as a promising further research. Another promising topic for a further analysis concerns extracting more relevant information from the databases, as the corresponding results showed that datasets that provide additional context or information layer to the data can help forecasting model improve their performance.

## 4   Conclusion

Empty container depots play an essential role in the empty container traffic and storage as well as in the inland transportation of goods. Due to unexpected or planned delays at ports or during transportation between port and the depot, certain setbacks can occur for the container transportation. Such incidents can create additional workload for the

Workload Forecasting of a Logistic Node Using Bayesian Neural Networks

empty container depot. Future workload forecast promises the potential for work optimization and efficient time management. Receiving the next week's forecast beforehand gives empty container depots the opportunity to preplan the work schedule and man power needed for the next week and enables them to redistribute the workforce of the depot when needed.

This paper has demonstrated that state-of-the-art Bayesian neural networks are capable of forecasting the hourly workload in an empty container depot for upcoming weeks in terms of the number of trucks handled at the depot and average handling time of the trucks.

This corresponds to a well-known problem of unpredictable workload changes at depots and offers a solution that addresses the problem with an automatable system. Moreover, it has shown that the forecasting model produces predictions that has a small error rate when more relevant data was provided, such as data from trucker appointment system of HCS-Depot.

Despite achieving a first step in forecasting hourly workload at empty container depots, it is clear that much remains to be done. Subsequent research will focus on introducing new relevant datasets, that the forecasting model can take advantage of, along with further developing the forecasting model. Accordingly, next steps will involve finding the relevant data sources and gathering the necessary datasets, applying further development work on the forecasting model as well as automatizing the process of publishing the forecast result. Since only aleatoric uncertainty was considered in this paper, including both aleatoric uncertainty and epistemic uncertainty in the model is one of the main milestones for the further development of the forecasting model. Using both uncertainties can help the model reduce predictive uncertainty, which is the confidence that a forecasting model has in a prediction. Moreover, since only certain parts of the databases were taken advantage of during forecasting experiments, it would be worthwhile to examine and work on these databases in detail to find out whether unused parts of the database can provide any further relevant information for better forecasting performance.

Workload Forecasting of a Logistic Node Using Bayesian Neural Networks

Workload Forecasting of a Logistic Node Using Bayesian Neural Networks